\documentclass[conference]{IEEEtran}
\IEEEoverridecommandlockouts
\usepackage{url}
\usepackage{cite}
\usepackage{amsmath,amssymb,amsfonts}
\usepackage{algorithmic}
\usepackage{graphicx}
\usepackage{textcomp}
\usepackage{xcolor}
\usepackage{graphicx}
\usepackage{caption}
\usepackage{lipsum}
\def\BibTeX{{\rm B\kern-.05em{\sc i\kern-.025em b}\kern-.08em
    T\kern-.1667em\lower.7ex\hbox{E}\kern-.125emX}}
\begin{document}

\title{Interactive Geometry Editing of Neural Radiance Fields
}

\author{\IEEEauthorblockN{Shaoxu Li}
\IEEEauthorblockA{\textit{John Hopcroft Center for Computer Science} \\
\textit{Shanghai Jiao Tong University}\\
Shanghai, China \\
lishaoxu@sjtu.edu.cn}
\and
\IEEEauthorblockN{Ye Pan*}
\IEEEauthorblockA{\textit{John Hopcroft Center for Computer Science} \\
\textit{Shanghai Jiao Tong University}\\
Shanghai, China \\
whitneypanye@sjtu.edu.cn}
}

\twocolumn[{%
\renewcommand\twocolumn[1][]{#1}%
\maketitle
\begin{center}
    \centering
    \captionsetup{type=figure}
    \includegraphics[width=0.8\textwidth]{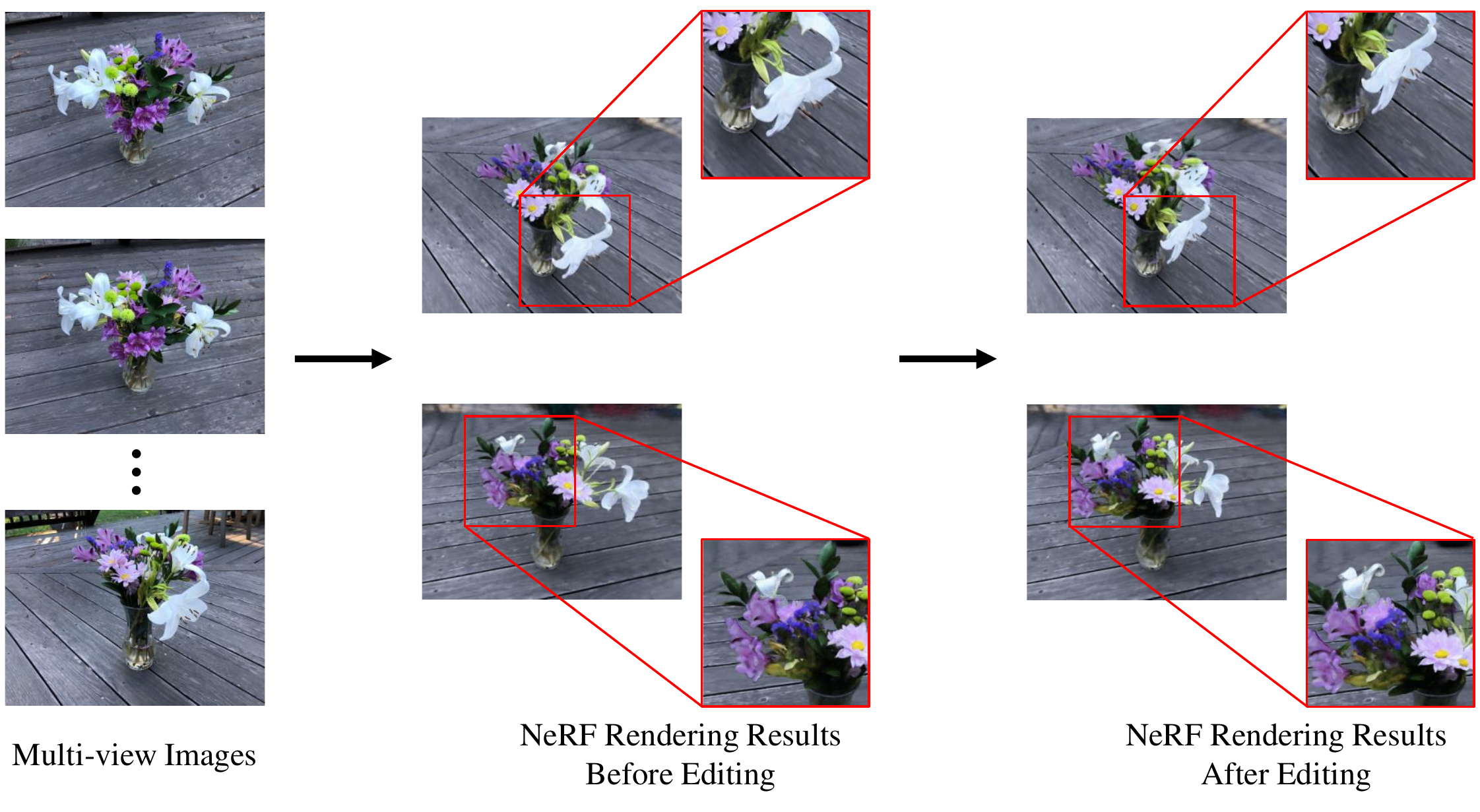}
    \captionof{figure}{Geometry editing results of our method. For a static neural radiance fields(NeRF), users can explicitly edit the implicit representation of the scene without leveraging explicit 3D representation. Editing adjusts the distribution of flowers while keeping multi-view consistency.}
\end{center}%
}]


\begin{abstract}
In this paper, we propose a method that enables interactive geometry editing for neural radiance fields manipulation. We use two proxy cages(inner cage and outer cage) to edit a scene. The inner cage defines the operation target, and the outer cage defines the adjustment space. Various operations apply to the two cages. After cage selection, operations on the inner cage lead to the desired transformation of the inner cage and adjustment of the outer cage. Users can edit the scene with translation, rotation, scaling, or combinations. The operations on the corners and edges of the cage are also supported. Our method does not need any explicit 3D geometry representations. The interactive geometry editing applies directly to the implicit neural radiance fields. Extensive experimental results demonstrate the effectiveness of our approach.
\end{abstract}


\section{Introduction}
Novel view synthesis has recently received increasing attention for its extensive application prospect, such as virtual and augmented reality, business, and artistic creation. Recently, neural radiance fields(NeRF)\cite{mildenhall2020nerf} has received increasing attention due to its high representation ability for 3D scenes. Fruitful follow-up works emerged for the improvement and application of NeRF, such as acceleration\cite{yu2022Plenoxels,Chen2022TensoRF,mueller2022instant}, scene generation\cite{Schwarz2020GRAF,Niemeyer2021GIRAFFE,Eric2021piGAN}, dynamic scene rendering\cite{park2021hypernerf,kania2022conerf}, scene style transfer\cite{Chiang2021Stylizing3S,Huang2022StylizedNeRF,chen2022upstnerf,li2023instant}, large-scale scene rendering\cite{Tancik2022Block,li2022read}. However, as an implicit 3D representation method, the neural radiance fields scenes are hard to edit and manipulate. Some researchers leveraged the explicit 3D representation to accomplish the editing task\cite{Yuan22NeRFEditing,xu2022deforming,peng2021CageNeRF,yang2022neumesh}, such as meshes. Users can intuitively edit the scene and accomplish shape deformation on the geometry. However, the transformation between NeRF and mesh is cumbersome, which impedes usability and quality. In this paper, we focus on the interactive geometry editing of neural radiance fields, with no need to acquire explicit 3D models.

3D shape editing is a popular research topic because of its applicability in games, movies, etc. Traditional 3D shape editing methods focus on explicit 3D models. Attribute-based model editing focus on the attributes of the surface model. The geometry characteristics and semantic attributes are the main constraints for model editing. Geometry-based deformation methods optimize the energy functions to accomplish the deformation problem, such as Laplacian-based mesh deformation\cite{kobbelt1998interactive}, as-rigid-as-possible\cite{alexa2000rigid}. Semantic constraints methods are high-level editing methods that edit multiple vertices simultaneously, such as FFD\cite{sederberg1986free}. Proxy-based deformation\cite{tagliasacchi20163d} uses a skeleton to guide the deformation of realistic animated characters. Cage-based deformation\cite{le2017interactive} uses cages wrapped outside the model for users' interactive editing. Data-Based deformation leverages shape datasets to produce more natural deformation results, such as Mesh-Based Inverse Kinematics\cite{sumner2005mesh}. Neural Shape Editing\cite{Tan_2018_Variational} has been popular in recent years. The latent feature provides a convenient way for shape editing.

The direct manipulation of NeRF for some specific tasks has been researched. Given a set of images for specific targets, such as human faces, 3D-aware image synthesis could be accomplished with generative methods, such as GRAF\cite{Schwarz2020GRAF}, GIRAFFE\cite{Niemeyer_2021_GIRAFFE}, pi-gan\cite{piGAN2021}. These methods can accomplish attribute editing and composition with 3D consistency for the target domain. Some methods use face parameters for human faces to embed the latent codes\cite{hong2021headnerf,Gafni_2021_Dynamic}. Controllable face generation represented by neural radiance fields can be accomplished. For video input, some methods\cite{park2021hypernerf} project NeRF into high-dimensional space for dynamic neural radiance fields modeling, enabling the deformation of neural radiance fields. For the editing of universal neural radiance fields, some methods leverage the explicit 3D representation to deform the scene. Existing researches\cite{Yuan22NeRFEditing,xu2022deforming,peng2021CageNeRF,yang2022neumesh,Jambon2023NeRFshop} extract meshes from the optimized radiance field, deform the meshes. And then, the deformation meshes guide the generation of a deformed radiance field. Although these methods above have made some achievements in radiance field manipulation, no methods can directly manipulate universal neural radiance fields scenes using interactive operations without explicit models.

We propose editing the neural radiance fields that interactively deforms the space to address the above issues. Unlike previous methods, our method does not need any explicit 3D geometry representations. Users can edit the neural radiance fields through direct interaction. We propose offering users two proxy cages for interactive geometry editing of the neural radiance fields. The two cages are inclusion relations. Users define the position of the two cages and manipulate the inner cage. The inner cage can be controlled with translation, rotation, scaling, and moving the corners or edges. The space inside the outer cage and outside the inner cage is adjusted with interpolation to promise the continuity of the scene. Users can accomplish simple deformation of the target scene. Complex deformation requires the users' multiple manipulations.

We conducted extensive experiments with synthetic datasets and natural datasets. Reasonable deformation and rendering results demonstrate the effectiveness of our method.

In summary, our contributions are listed as follows:
\begin{itemize} 
\item We propose the first method to support interactive controlling of the shape deformation of neural radiance fields(NeRF) without mesh extraction.

\item We designed an interactive mode with two proxy cages and various operations for interactive geometry editing of radiance fields.

\item We conducted extensive experiments to execute the operations we designed, and the rendering results demonstrate the effectiveness of our approach.
\end{itemize}

\section{Related Work}
\noindent\textbf{Novel View Synthesis.}
Novel view synthesis is a significant task in computer vision. Traditional methods, such as meshes and point clouds, use explicit representation methods for 3D object or scene modeling. Recently, neural rendering has received increasing attention. The neural radiance fields(NeRF)\cite{mildenhall2020nerf} is the representative work, which uses a multilayer perceptron (MLP) to model the scene. 
The model learns the target scene from a set of images and outputs photo-realistic rendering with volume rendering. Many follow-up works emerged for higher quality, faster speed, and more application. Although these works improve the quality and practicability of NeRF, editing of NeRF is still unexplored due to its implicit method.

\noindent\textbf{Explicit 3D Geometry Deformation.}
Interactive geometry editing of the 3D object is vital for artistic creation and production. For interactive geometry editing of explicit 3D models, many mature algorithms have been widely used in commercial software. Free-Form Deformation(FFD)\cite{sederberg1986free} is a primary mesh deformation method that embeds the 3D model into the deformation space. Then, the deformation space is manipulated to deform the embedded geometric model. FFD can not control large meshes and does not consider object morphology. Cage-based deformation(CBD)\cite{Joshi2007Harmonic,YifanNeuralCage2020} manipulates the vertices surrounding the object cage to deform the object. The new position of mesh points can be calculated from the relative transformation between the cage and deformed cage coordinates. In our method, users use two nested cages for shape deformation. The control accuracy of our method is between the FFD\cite{sederberg1986free} and CBD\cite{Joshi2007Harmonic,YifanNeuralCage2020} for meshes.

\begin{figure*}[ht]
\centering
\includegraphics[width=\linewidth]{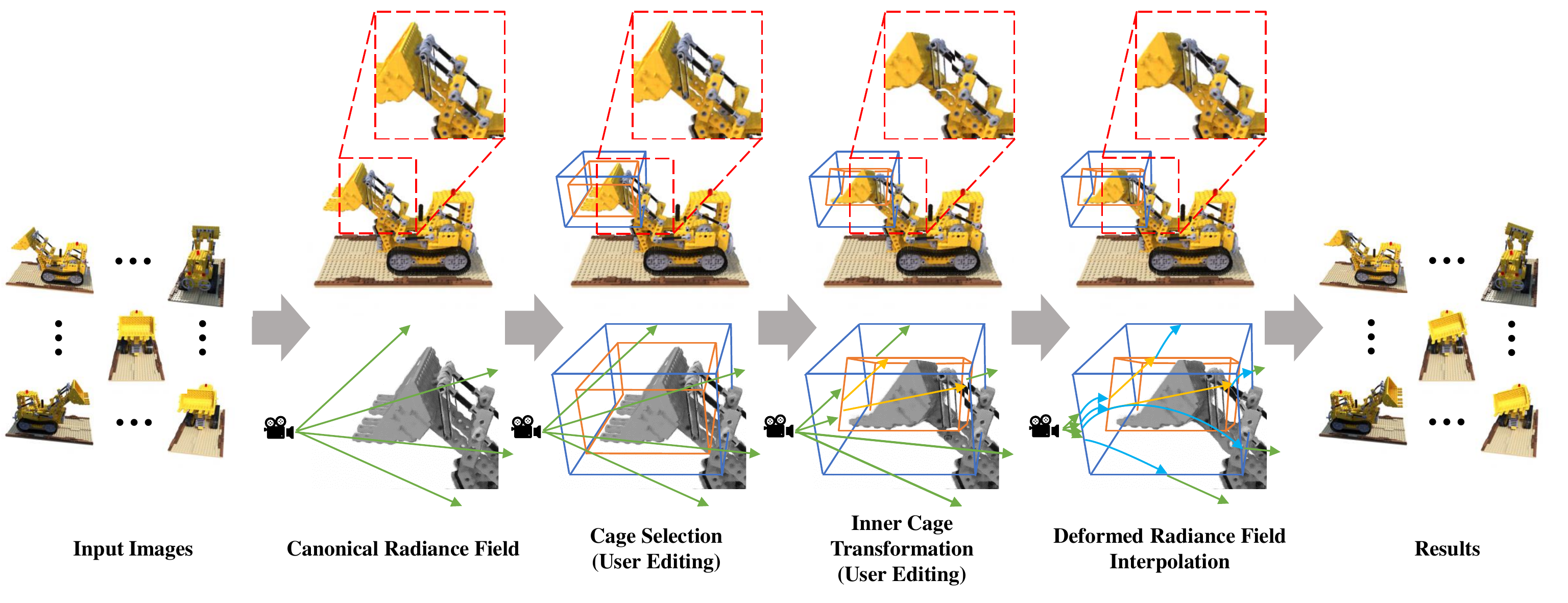}
\caption{The architecture of our method. Our approach uses two proxy cages to execute interactive neural radiance fields editing. Users set two cages first. Then users can edit the inner cage with predefined operations, including cage transformation(translation, rotation,
scaling) and cage deformation(moving the corners or edges of
the cage). The radiance field inside the inner cage change with the cage. Two kinds of deformation can be chosen for the deforming of radiance fields. The discrete adjustment only adjusts the inner cage areas. Continuous adjustment uses interpolation to promise the continuity of the radiance field.}
\label{fig:model}
\end{figure*}
\noindent\textbf{Neural Radiance Fields Deformation.} The editing of a universal static NeRF scene is still unexplored. NeRF-Editing\cite{Yuan22NeRFEditing} extracts an explicit triangular mesh representation from the
trained NeRF, and the explicit mesh representation is then intuitively deformed by the user. The mesh deformation propagates the scene's geometric surface deformation to the spatial discrete deformation field. DRFC\cite{xu2022deforming} use a triangular mesh that encloses the foreground object called cage as an interface. By manipulating the cage vertices, this approach enables the free-form deformation of the radiance field. CageNeRF\cite{peng2021CageNeRF} performs the deformation on an enclosing polygon mesh with sparsely defined vertices called cage inside the rendering space, where each point is projected into a novel position based on the barycentric interpolation of the deformed cage vertices. NeuMesh\cite{yang2022neumesh} proposed to encode the implicit field into mesh vertices. By simply deforming the corresponding mesh, the change will synchronously take effect on the implicit field, and the rendered object will also be deformed accordingly. NeRFshop\cite{Jambon2023NeRFshop} provides a scribble-based interface for interactive object selection in NeRFs and semi-automatic cage building. The user scribbles on the target object or volumetric region in space. NeRFshop reprojects the scribbles into the 3D volume, performs region growing, builds a cage, and discretizes the volume using a tetrahedral mesh. Manipulating the corners of the cages accomplishes free-form editing of NeRF. All these methods with meshes above can edit NeRF with good quality. The meshes increase the complexity and time-consuming of editing. We propose a novel interactive geometry editing method without mesh, which is simpler and faster.

\section{Method}
We deform the NeRF scene by manipulating the space directly. As shown in Fig.\ref{fig:model}, the first step is to select two proxy cages for deformation, the inner cage for interactive manipulation and the outer for automatic adjustment. The inner cage is inside the outer cage. Then, the user can manipulate the inner cage optionally, including translation, rotation, scaling, shearing, or any combination of these. And the movement of corners and edges of the inner cage is also allowed. The scenes are updated with the manipulation of users. The radiance field inside the inner cage change with the cage. Two kinds of deformation can be chosen for the deforming of radiance fields. The discrete adjustment only adjusts the inner cage areas. Continuous adjustment uses interpolation to promise the continuity of the radiance field. Other regions inside the outer cage are adjusted using an interpolation algorithm. The regions outside of the outer cage remain unchanged.

\subsection{Neural Radiance fields}
We first use a typical NeRF to optimize the radiance field for the scene to be edited. Neural Radiance fields(NeRF) proposes using a multi-layer perceptron (MLP) to model a scene, given a set of images from different views. NeRF generates light rays from image pixels toward the scene, given camera parameters and positions. For the rays, NeRF sample points on the ray and embeds the position of the point $\bold{p}\in \mathbb{R}^3$ and ray direction $\bold{d}\in \mathbb{R}^3$ for volume rendering. The networks learn the mapping from embed features to the RGB color $\bold{c}\in \mathbb{R}^3$ and density $\sigma \in \mathbb{R}^3$ of the points:
\begin{equation}
    \bold{F}_\Theta : (\gamma(\bold{p}),\gamma(\bold{d})) \rightarrow (\bold{c},\sigma)
\end{equation}
where $\gamma(\cdot)$ is the position embedding function, $\Theta$ is the trainable parameters. Given the ray $\bold r$ from image pixels, photorealistic rendering can be accomplished with the color and density using discrete integration.

Since the rendering is based on sampled points, adjusting the points' coordinates acts on the rendering results. For NeRF manipulation, deforming the coordinates space is workable for deforming the NeRF scene.

\subsection{Interactive Operations} \label{Interactive Operations}
To interactively deform the neural radiance fields, we propose explicitly manipulating the radiance fields without any explicit 3D models. We manipulate with two cages in space, one inner cage, and one outer cage. The inner cage is inside the outer cage. The two cages are manually set, and the inner cages can be moved or edited. When the inner cage is modified, the space inside the outer cage is adjusted accordingly. The space outside the outer cage remains unchanged. The inner cage executes the deformation intentions of users. Meanwhile, the outer cage promises the continuity of the radiance field scene when using continuous adjustment.

\subsubsection{Cage Setting}
\label{Cage setting}
\begin{figure}[ht]
\centering
\includegraphics[width=0.6\linewidth]{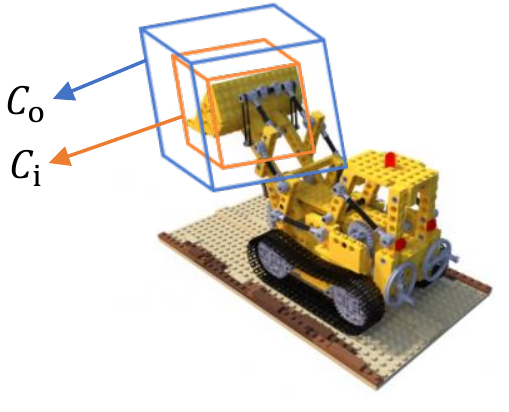}
\caption{Users shall define two cages manually. The inner cage is inside the outer. In the cage setting process, the radiance field remains unchanged.}
\label{fig:two_cages}
\end{figure}

For interactive editing, users shall define two cages manually to confine the operation area. As shown in Fig.\ref{fig:two_cages}, two cages wrap part of the radiance field. The edges of cages are rendered together with the neural radiance fields. For later calculation, the two cages are denoted as $\boldsymbol{C}_{i}$ with vertices ${\bold{v}_{i}}$, $\boldsymbol{C}_{o}$ with vertices ${\bold{v}_{o}}$. All the transformation operations in Sec.\ref{Cage Transformation} are supported in the cage setting process. The setting operations only transform the shape and position of the two cages, leaving the radiance field unchanged. The outer cage can be moved in the whole radiance field space and confines the motion of the inner cage. In our method, we use cube cages for illustration. For free editing, the cages are not restricted to typical shapes. Users can even indicate some control points for NeRF deformation. The core idea of our method is to define two spaces to deform and adjust the scene.
\subsubsection{Cage Transformation}
\label{Cage Transformation}
Users can operate two cages with different operations, including translation, rotation, scaling, or any combinations. Since these operations shall take the cage center as the central, the coordinates shall be adjusted for transformation. The center changes with translation, rotation, and scaling operations for simplification. The cage center remains unchanged for the cage deformation operations in Sec.\ref{Cage Deformation}. In this instance, the point coordinates adjustment is easy to implement. Given a cage $\boldsymbol{C}$ with vertices ${\bold{v}_{j}}$, points $\bold{p}\in \mathbb{R}^3$ in space can be adjusted to $\bold{p}_{c}\in \mathbb{R}^3$:

\begin{equation}
\label{equation:center adjust}
\bold{p}_{c}=\bold{p}-\frac{1}{6}\sum_{j} \bold{v}_{j}
\end{equation}
After adjustment, the translation, rotation, and scaling could be accomplished through user interactions. We capture the user actions into transformation parameters, including translation vector $(t_x,t_y,t_z)$, rotation vector $(\theta_x,\theta_y,\theta_z)$, and scale vector $(s_x,s_y,s_z)$. $(t_x,t_y,t_z)$ denotes the translation along the $X,Y,Z$ axes. $(\theta_x,\theta_y,\theta_z)$ denotes the rotation angles along the $X,Y,Z$ axes. $(s_x,s_y,s_z)$ denotes the scale factor along the $X,Y,Z$ axes. For points $\bold{p}=(x,y,z)$ inside the cage, the adjusted coordinates are denoted as $\bold{p}_{c}=(x_c,y_c,z_c)$. The transformation is accomplished using Eq.\ref{equation:Transformation}.
\begin{equation}
\label{equation:Transformation}
\begin{bmatrix}
x_c'\\y_c'\\z_c'\\1
\end{bmatrix}
=
T R_x R_y R_z S
\begin{bmatrix}
x_c\\y_c\\z_c\\1
\end{bmatrix}
\end{equation}
where $(x_c',y_c',z_c')$ denotes the transformed coordinates, $T$ denotes the translation matrix, $R_x R_y R_z$ denote the rotation matrices, $S$ denote the scale matrix. These matrices are calculated using Eq.\ref{equation:translation},\ref{equation:rotation},\ref{equation:scaling}.
\begin{equation}
\label{equation:translation}
T=
\begin{bmatrix}
1&0&0&t_{x}\\
0&1&0&t_{y}\\
0&0&1&t_{z}\\
0&0&0&1
\end{bmatrix}
\end{equation}
\begin{equation}
\label{equation:rotation}
\begin{split}
R_x=
\begin{bmatrix}
1&0&0&0\\
0&cos\theta_x&-sin\theta_x&0\\
0&sin\theta_x&cos\theta_x&0\\
0&0&0&1
\end{bmatrix}\\
R_y=
\begin{bmatrix}
cos\theta_y&0&sin\theta_y&0\\
0&1&0&0\\
-sin\theta_y&0&cos\theta_y&0\\
0&0&0&1
\end{bmatrix}\\
R_z=
\begin{bmatrix}
cos\theta_z&sin\theta_z&0&0\\
-sin\theta&cos\theta_z&0&0\\
0&0&1&0\\
0&0&0&1
\end{bmatrix}
\end{split}
\end{equation}

\begin{equation}
\label{equation:scaling}
S=
\begin{bmatrix}
s_x&0&0&0\\
0&s_y&0&0\\
0&0&s_z&0\\
0&0&0&1
\end{bmatrix}
\end{equation}

\subsubsection{Cage Deformation}
\label{Cage Deformation}
Besides transforming the whole cage, the deformation of cages is also available. Users can deform the inner or outer cage by moving the corners or edges of the cages. Given a cage $\boldsymbol{C}$ with vertices ${\bold{v}_{j}}$, edges ${\bold{e}_{k}}$, users can drag any visible vertices or edges to available positions. After deformation, the new vertices ${\bold{v}_{j}'}$ and edges ${\bold{e}_{k}'}$ constitute the new cages. 

The cage transformation does not change the shape of the two cages. The cage deformation manipulates the shape of the two cages. In the cage transformation, the coordinates transformation of points inside the cage is the same as the cage corners. In the cage deformation, the points' coordinates inside the cage are calculated with interpolation using the cage corners coordinates.

\begin{figure*}[t]
\centering
\includegraphics[width=\linewidth]{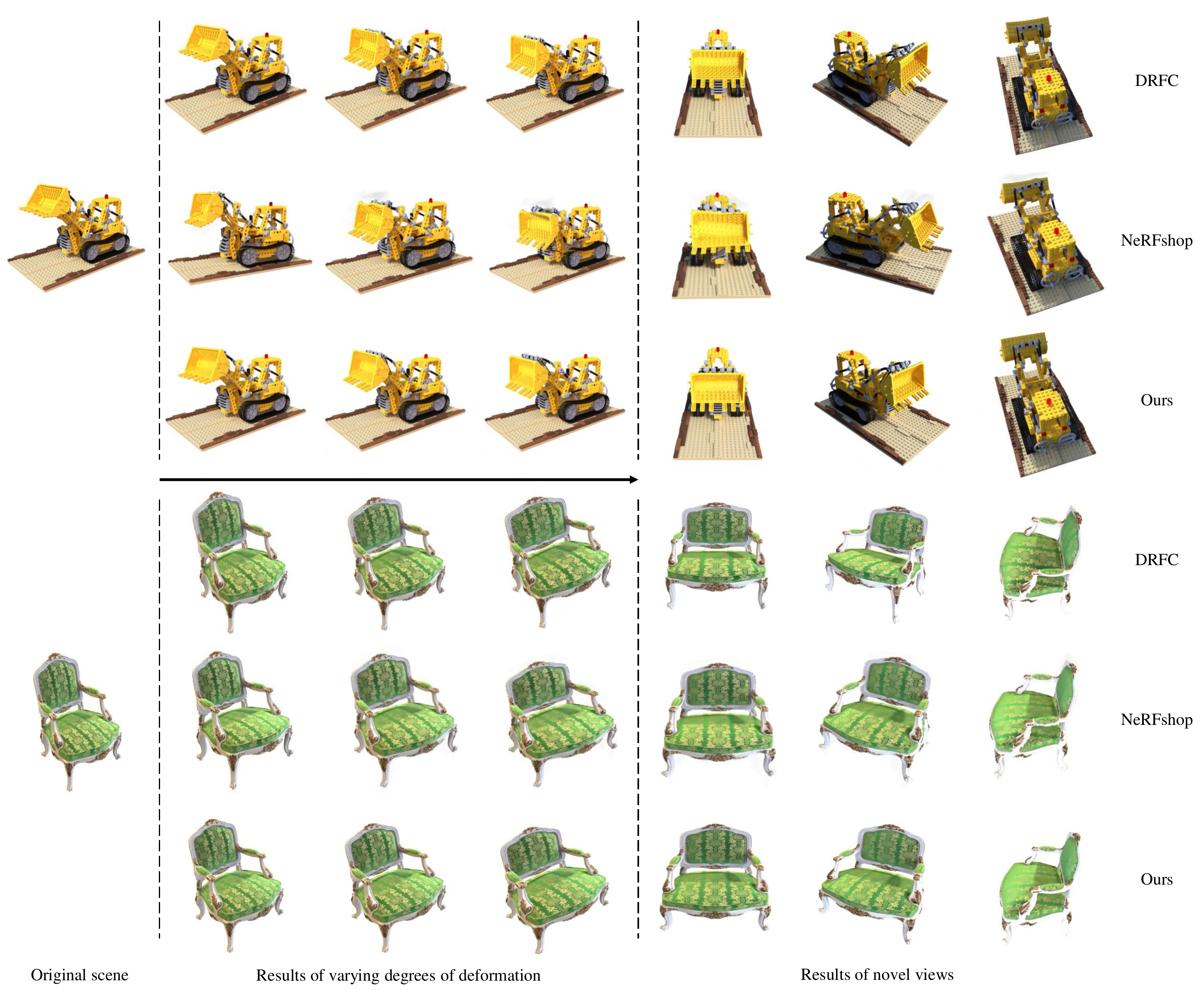}
\caption{Comparison with state-of-the-art methods. From left to right: original scenes, results of varying degrees of deformation, results of novel views. From top to bottom(each scene), DRFC, NeRFshop, and ours.}
\label{fig:comparison}
\end{figure*}

\begin{figure*}[t]
\centering
\includegraphics[width=\linewidth]{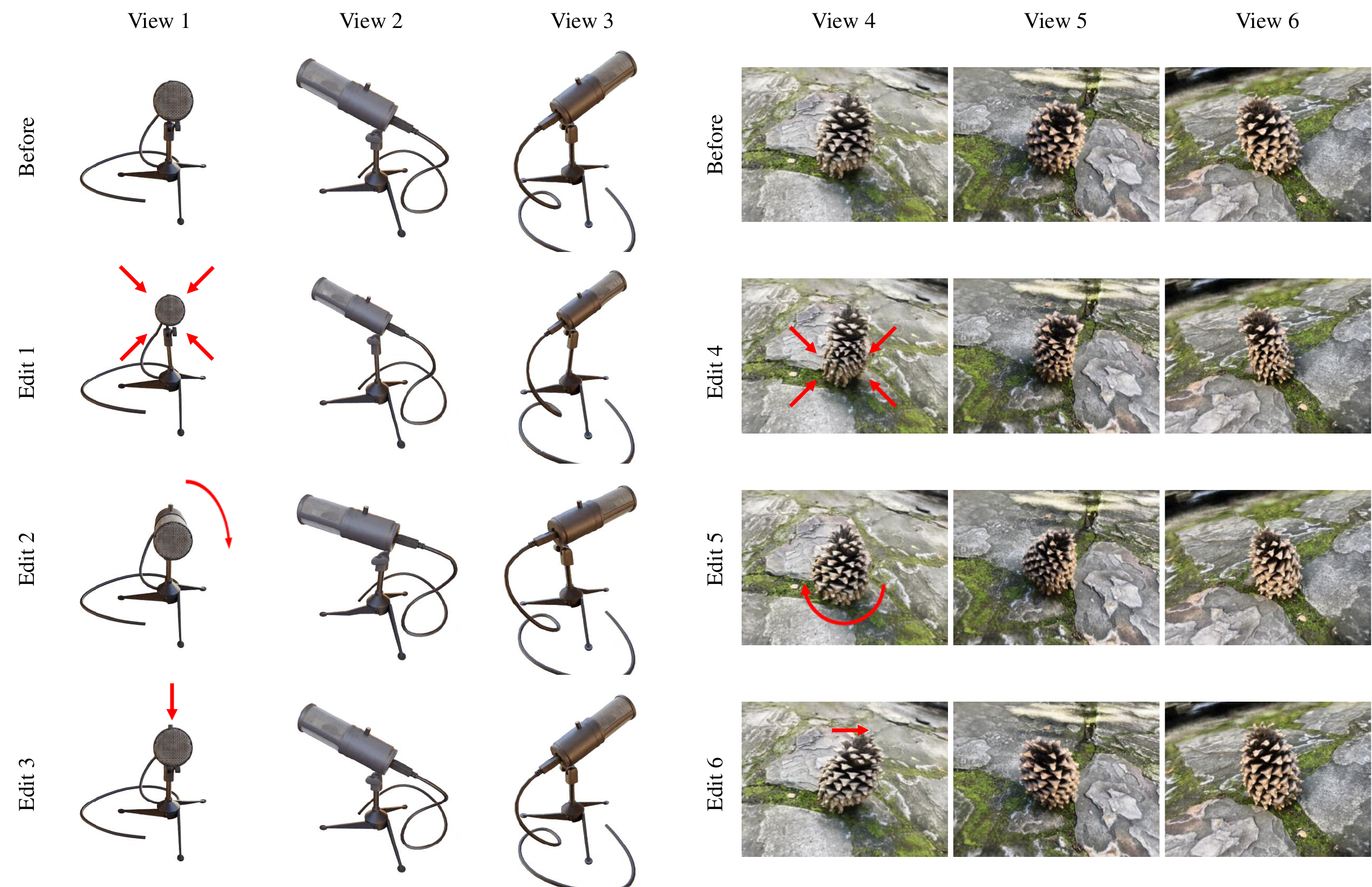}
\caption{Results of different edits. The mic is a synthesis scene. The pinecone is a natural scene. The red arrows indicate the editing details. We show the editing results from three views for each edit of two scenes.}
\label{fig:operations}
\end{figure*}

\subsection{Deforming Radiance Fields}
In our method, users use two cages to deform the radiance fields, the inner and outer cages. Users can manipulate the two cages with all the interactive operations in Sec.\ref{Interactive Operations}. Only the cage moves for the cage setting process in Sec.\ref{Cage setting}. In this section, the space is warped to achieve the desired editing for the deformation of radiance fields.

The deformed radiance field $\Psi$ can be divided into four parts depending on the space that: (1) outside the outer cage, (2) inside the outer cage but outside the canonical inner cage and transformed inner cage, (3) inside the canonical cage but outside the transformed inner cage, (4) inside the transformed inner cage. When editing the inner cage, two kinds of deformation can be chosen, discrete and continuous adjustment.

\subsubsection{Discrete Adjustment}
For discrete adjustment, the operation on the inner cage only changes the rendering results inside the inner cage. We denote the transformed-to-canonical mapping of spatial position and view direction as:
\begin{equation}
    \phi_\bold{p}:\bold{p}\rightarrow \bold{p}^c, \phi_\bold{d}:(\bold{p},\bold{d})\rightarrow \bold{d}^c, 
    \label{eq:mapping1}
\end{equation}
where $\bold{p}^c$ can be obtained form operations in Sec.\ref{Cage Transformation} and \ref{Cage Deformation}. For cage transformation, the mapping is the transformation in Eq.\ref{equation:Transformation}. For cage deformation, the mapping is obtained with trilinear interpolation, whose control points are sampled from the corners of the inner cage.

Given outer cage $\boldsymbol{C}_o$, canonical inner cage $\boldsymbol{C}_i^c$, transformed inner cage $\boldsymbol{C}_i$, the transformation function can be obtained using Eq.\ref{eq:Discrete adjustment}.
\begin{equation}
    \Psi^D(\bold{p},\bold{d})=
        \begin{cases}
        \Psi(\bold{p},\bold{d}), &\bold{p} \in \mathbb R^3 \backslash (\mathbb V^c \cup \mathbb V)\\
        (\textbf{0},0) \; or \; \Psi(\bold{p},\bold{d}), &\bold{p} \in \mathbb V^c \backslash (\mathbb V^c \cap \mathbb V)\\
       \Psi(\phi_\bold{p}(\bold{p}),\phi_\bold{d}(\bold{p},\bold{d})), &\bold{p} \in  \mathbb V\\
        \end{cases}
    \label{eq:Discrete adjustment}
\end{equation}
where $\mathbb V^o,\mathbb V^c,\mathbb V \in \mathbb R^3$ denotes the space enclosed by $\boldsymbol{C}_o$, $\boldsymbol{C}_i^c$ and $\boldsymbol{C}_i$, respectively. The space inside the inner cage in the deformed radiance field is substituted by the space inside the inner cage in the canonical radiance field. For the areas the inner cage used to be, empty rendering results(zero density) are filled for object movement or deletion. It can also remain unchanged for object copy. The space outside the inner cage remains unchanged. In this way, the movement of the operation target is independent of the whole radiance field, which means disconnection may emerge with the operation.

\subsubsection{Continuous Adjustment}
For continuous adjustment, the operation with the inner cage changes the rendering results inside the outer cage. The transformation inside the inner cage is obtained using Eq. \ref{eq:mapping1}. The transformation outside the inner cage while inside the outer cage is obtained with trilinear interpolation, whose control points are sampled from the surface of cages. We denote the transformed-to-canonical mapping of spatial position and view direction in space $\bold{p} \in \mathbb V^o \backslash (\mathbb V^c \cup \mathbb V)$ as:
\begin{equation}
    \phi'_\bold{p}:\bold{p}\rightarrow \bold{p}^c, \phi'_\bold{d}:(\bold{p},\bold{d})\rightarrow \bold{d}^c, 
    \label{eq:mapping2}
\end{equation}

Given outer cage $\boldsymbol{C}_o$, canonical inner cage $\boldsymbol{C}_i^c$, transformed inner cage $\boldsymbol{C}_i$, the transformation function can be obtained using Eq.\ref{eq:Continuous adjustment}.
\begin{equation}
    \Psi^D(\bold{p},\bold{d})=
        \begin{cases}
        \Psi(\bold{p},\bold{d}), &\bold{p} \in \mathbb R^3\backslash \mathbb V^o  \\ 
        \Psi(\phi'_\bold{p}(\bold{p}), \phi'_\bold{d}(\bold{p},\bold{d})), &\bold{p} \in \mathbb V^o \backslash \mathbb V\\
       \Psi(\phi_\bold{p}(\bold{p}), \phi_\bold{d}(\bold{p},\bold{d})), &\bold{p} \in  \mathbb V\\
        \end{cases}
    \label{eq:Continuous adjustment}
\end{equation}
The space inside the inner cage in the deformed radiance field is substituted by the space inside the inner cage in the canonical radiance field. For the areas outside the inner cage but inside the outer cage, interpolation results are calculated according to the coordinates deformation in the surface of the inner and outer cage. The space outside the inner cage remains unchanged. In this way, the continuity of the radiance field remains unchanged before and after the deformation.

\begin{figure*}[t]
\centering
\includegraphics[width=\linewidth]{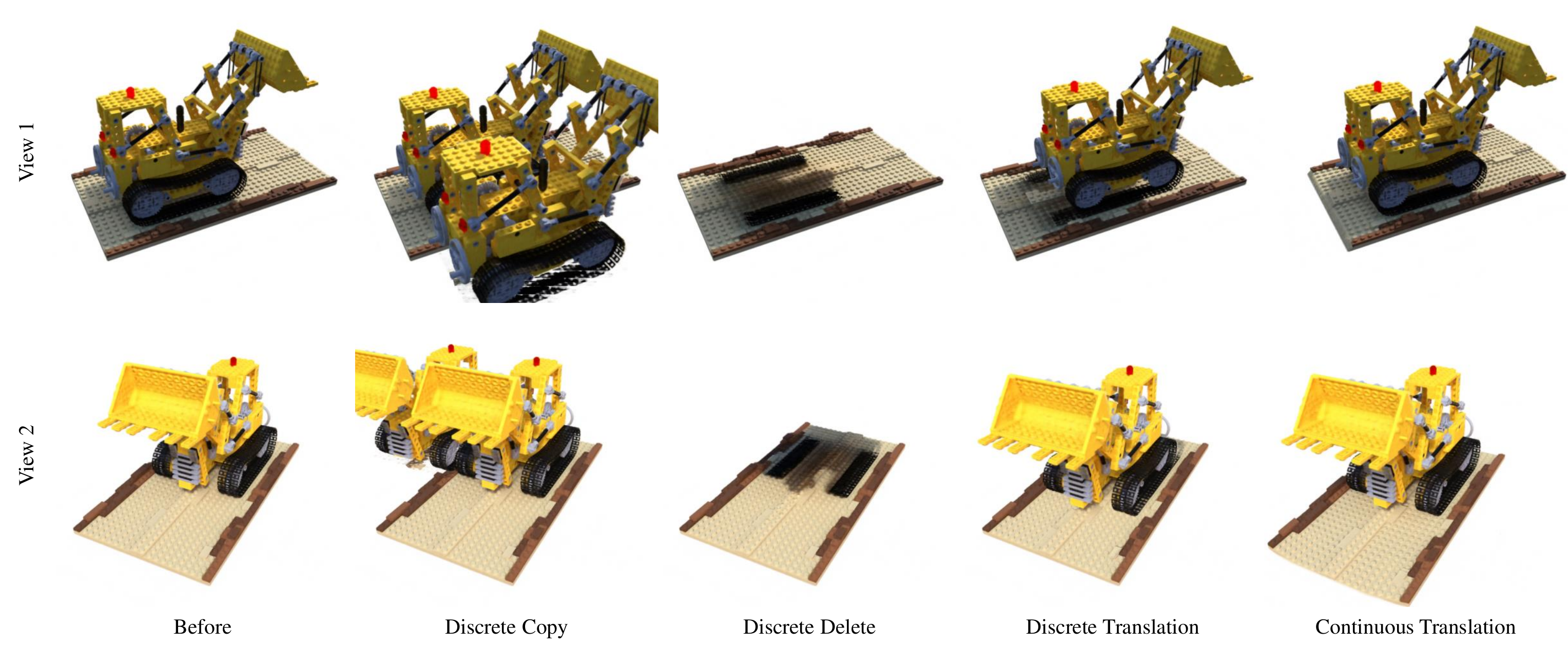}
\caption{Results of discrete adjustment. The Lego loader is edited with copy, delete, and translation. And we show the comparison with continuous adjustment.}
\label{fig:result_discrete}
\end{figure*}

\section{Experiments}
\label{sec:experiments}
In this section, we evaluate our approach's effectiveness through various scenes. We show the results of extensive ablation studies and then discuss the limitations of our approach. Our results are based on the original NeRF\cite{mildenhall2020nerf} implemented by NeRF-pytorch\cite{lin2020nerfpytorch}. Since we only manipulate the points coordinates, our method applies to various modified NeRF methods. The code is accomplished using Pytorch on a single Nvidia 3080 GPU.
\subsection{Comparison}
For arbitrary scene editing, the ground truth results of the deformation are not available. For different NeRF editing methods, it is hard for users to accomplish duplicate operations. Considering this, we execute comparisons on NeRF-Synthetic dataset\cite{mildenhall2020nerf} for basic editing operations. We compare our method with state-of-the-art methods DRFC\cite{xu2022deforming} and NeRFshop\cite{Jambon2023NeRFshop}. Similar editing operations are executed on the radiance fields. Our method directly interacts with the radiance field, while DRFC and NeRFshop extract meshes, edit the meshes cages, and then deform the radiance fields. Fig.\ref{fig:comparison} shows the results. Two scenes are edited respectively. In the original scene, the results of varying degrees of deformation and novel views of edited results are shown from left to right. The shovel moves from top to bottom for the Lego and rotates with an angle. For the chair, the height decreases while the width increases.

The editing methods with mesh cages manipulate the scene with cage deformation. When moving the cage corners, these corners and the space inside each tetrahedron are adjusted. Since the cage is generated with algorithms, the adjustment is not always ideal. Our method maintains the space inside the inner cage relatively constant. As shown in the Lego results of DRFC and NeRFshop, different degrees of distortion has occurred with the shovel shape. In our results for Lego, the shape changes only with the joint space, and the shovel and loader body remain unchanged. Our method and DRFC are based on the original NeRF. The NeRFshop is based on the Instant-NGP, which uses a hash table to encode position. The method uses occupancy grids to represent the scene and skips the empty areas in optimization. For the empty space in the results of NeRFshop, color, and density artifacts exist because of the occupancy grids, as shown in Fig.\ref{fig:comparison}.

\subsection{Operations}
To illustrate the effectiveness of our method, we manually edit the scene with all the operations we define. Two kinds of operations are supported in our method. The first is cage transformation, and the second is cage deformation. Fig.\ref{fig:operations} shows the novel view synthesis results of the original and deformed scene. Two scenes from NeRF-Synthetic\cite{mildenhall2020nerf} and Nerf360\cite{barron2021mipnerf} are optimized and edited manually with different operations. The mic scene is a rendered scene from the NeRF-Synthetic dataset. The first line shows the rendering results of different views. The second to fourth lines show the editing results of scaling, rotation, and translation of the receiver of the mic separately. The pinecone scene is a natural scene from Nerf360 dataset. The first line shows the rendering images of different views. The second to fourth lines show the editing results of lower half scaling with top unchanged, top rotation with bottom unchanged, and top translation with bottom unchanged of the pinecone.

\subsection{Ablation Study}
\begin{figure*}[t]
\centering
\includegraphics[width=1.0\linewidth]{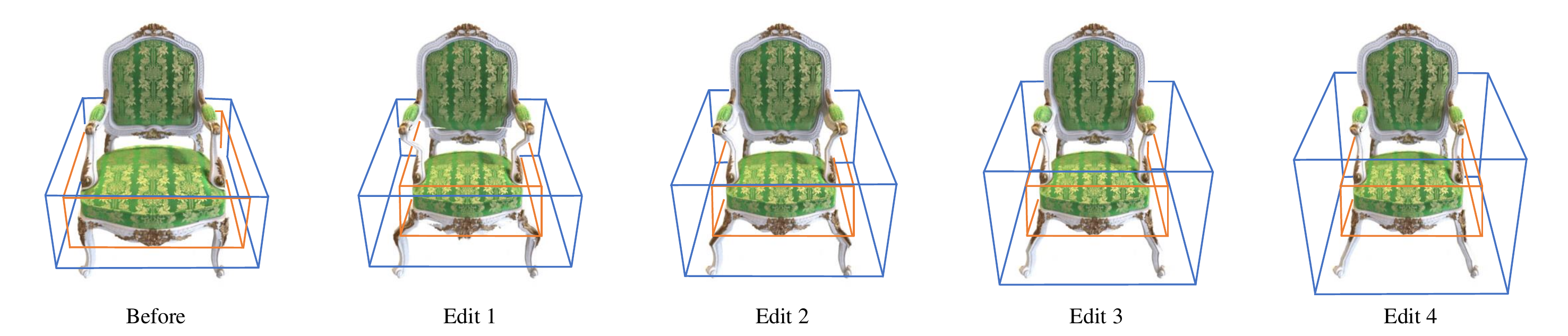}
\caption{Impact of the relative size of the inner and outer cage. For a fixed inner cage, the synthesis results get smoother with the increase of the outer cage. A small outer cage may lead to discontinuity.}
\label{fig:cage_size_result}
\end{figure*}

\subsection{Discrete Adjustment}
In the cases above, we show the editing results with various operations on the synthesis and natural scenes. All the results above are continuous adjustment, which adjusts the space outside the inner cage and inside the outer cage for continuity between the inner cage and the space outside the outer cage. Our method also supports discrete adjustment. For discrete adjustment, the manipulation only applies to the inner space. Besides the simple deformation, object copy and deletion are also achievable. Fig.\ref{fig:result_discrete} shows the comparison between continuous adjustment and discrete adjustment. The user manipulates the Lego loader with a translation. The inner cage surrounds the loader. The outer cage surrounds the loader and floor. Copy operation creates another loader. Delete operation removes the loader. Discrete translation translates the loader with a distance. These discrete operations do not influence the floor. Continuous translation translates the loader with the same distance, while the floor has apparent displacement and deformation. 

Ideal results can be achieved with discrete adjustment. However, it is hard to separate the coterminous targets. The results in Fig.\ref{fig:result_discrete} illustrate the problem. The boundary between the floor and the loader adheres to each other in the copy results.

\noindent\textbf{Impact of outer cage size.}
In our method, the inner cage defines the space to be manipulated, and the outer cage defines the space to be adjusted. When editing the inner cage, the adjustment varies with the relative size of the inner and outer cages. As shown in Fig.\ref{fig:cage_size_result}, we edit the inner cage with the same scaling operation. The outer cages get bigger and bigger from left to right. With the increase of the outer cage, the adjusted space gets bigger, and the synthesis result gets smoother. The results with the smallest outer cage have discontinuities. It is essential to select a proper cage size.

\begin{figure}[h]
\centering
\includegraphics[width=\linewidth]{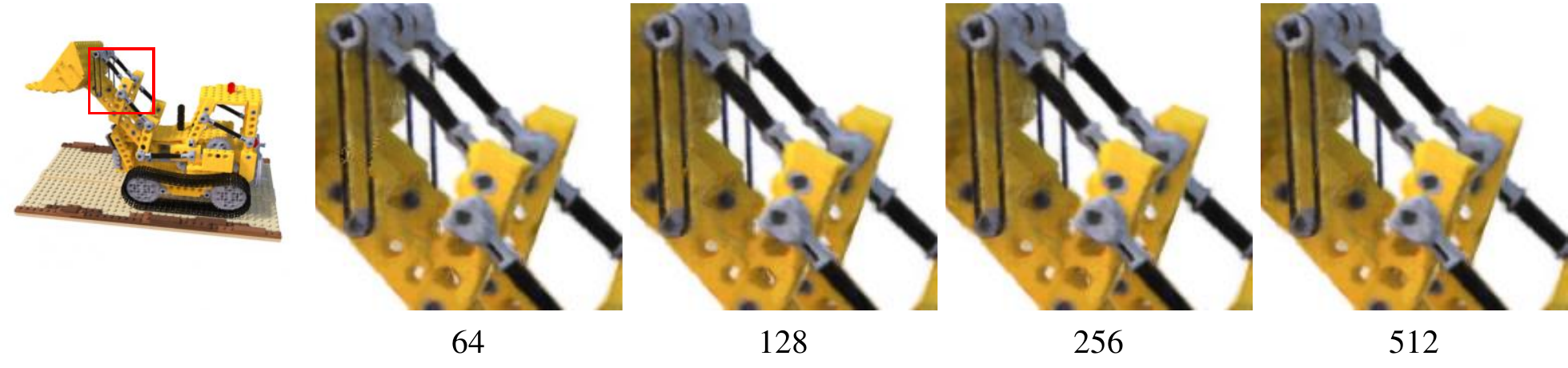}
\caption{Impact of discretization resolution. The synthesis quality increases with the resolution. Low resolution may lead to discontinuities.}
\label{fig:resolution_result}
\end{figure}

\noindent\textbf{Impact of discretization resolution.}
The mapping from canonical-to-deformed stretches or compresses the space. High resolution for a typical NeRF leads to good rendering quality while increasing the computation time and space. Fig.\ref{fig:resolution_result} shows the synthesized results with different resolutions. For 64 resolution, significant discontinuities can be observed. With the increase in resolution, the discontinuities decrease. For 256 resolution, the discontinuities can hardly be observed. Our results in Sec.\ref{sec:experiments} are accomplished with resolution 256.

\section{Disscussion and Limitations}

\subsection{Disscussion} We propose the first interactive editing method for neural radiance fields without mesh extraction. Compared with state-of-art NeRF editing methods with automatic mesh cage generation, we propose to allow users to define simple cages freely. Although automatic mesh cage generation makes user interaction easier, the generated cages are unfixed. The cages may be different in two cases for the same scene.

\subsection{Limitations}There are some limitations to our method. Firstly, reasonable canonical radiance field space is desirable for deformation. In our method, the deformation results of the space inside the outer cage and outside the inner cage are an interpolation. For some scenes with images from a single side, parts of the scene are under-modeled due to occlusion. The modeling results are not matched with reality space, although the rendering in some views is satisfactory. Unideal canonical radiance field space may lead to lousy deformation results. 

Secondly, we use two cages to accomplish the manipulation of space. Two cages are far from conforming to the shape details of a scene. That means the manipulation of our method is coarse. For delicate manipulation, users need to do multiple operations.

\begin{figure}[h]
\centering
\includegraphics[width=0.6\linewidth]{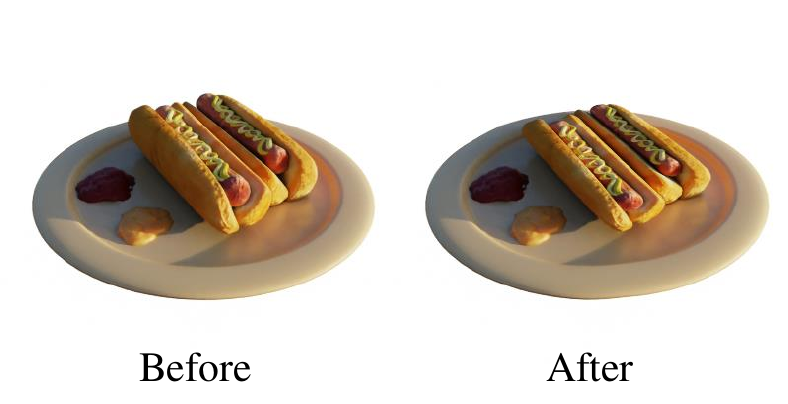}
\caption{When adjusting part of the scene, the other space remains unchanged.}
\label{fig:illu}
\end{figure}

Thirdly, deforming the space does not maintain the consistency of shadow and light, as shown in Fig.\ref{fig:illu}. The space is only adjusted according to the shape when editing the target. NeRF scenes store the shadow and light as the texture characteristics of the space. The editing with the consistency of shadow and light shall be solved in future research.

\section{Conclusion}
In this paper, we propose the first method to support interactive controlling of the shape deformation of neural radiance fields(NeRF) without mesh extraction. Users can interactively edit the radiance field with various operations by choosing the deformation area(inner cage) and the adjustment area(outer cage) with two cages. Predefined operations include cage transformation(translation, rotation, scaling on the whole cage) and deformation(movement on the cage corners or edges). Discrete and continuous adjustments can be chosen. The discrete adjustment only moves the deformation target, while continuous adjustment adjusts the outer space for continuity. We demonstrate the effectiveness of our method on synthetic and natural scenes, comparing it with state-of-art methods. Our method is accomplished with the original NeRF, which applies to various modified NeRF methods.
\bibliographystyle{IEEEbib}
\bibliography{icme2023template}

\end{document}